\documentclass[11pt,a4paper]{article}
\usepackage{authblk}
\usepackage[hyperref]{acl2017}
\usepackage{times}
\usepackage{latexsym}

\usepackage{url}

\usepackage[T1]{fontenc}
\usepackage{amssymb}
\usepackage{amsmath}
\usepackage{graphicx}
\usepackage{subcaption}

\aclfinalcopy 

\title{Knowledge Adaptation: Teaching to Adapt}

\author[1,2]{\bf Sebastian Ruder}
\author[2]{\bf Parsa Ghaffari}
\author[1]{\bf John G. Breslin}
\affil[1]{Insight Centre for Data Analytics}
\affil[ ]{National University of Ireland, Galway}
\affil[2]{Aylien Ltd.}
\affil[ ]{Dublin, Ireland}
\affil[ ]{\tt \{sebastian.ruder,john.breslin\}@insight-centre.org}
\affil[ ]{\tt \{sebastian,parsa\}@aylien.com}

\date{}

\begin{document}
\maketitle\begin{abstract}
Domain adaptation is crucial in many real-world applications where the distribution of the training data differs from the distribution of the test data. Previous Deep Learning-based approaches to domain adaptation need to be trained jointly on source and target domain data and are therefore unappealing in scenarios where models need to be adapted to a large number of domains or where a domain is evolving, e.g. spam detection where attackers continuously change their tactics.

To fill this gap, we propose \emph{Knowledge Adaptation}, an extension of \emph{Knowledge Distillation} \cite{Bucilua2006, Hinton2015} to the domain adaptation scenario. We show how a student model achieves state-of-the-art results on unsupervised domain adaptation from multiple sources on a standard sentiment analysis benchmark by taking into account the domain-specific expertise of multiple teachers and the similarities between their domains.

When learning from a single teacher, using domain similarity to gauge trustworthiness is inadequate. To this end, we propose a simple metric that correlates well with the teacher's accuracy in the target domain. We demonstrate that incorporating high-confidence examples selected by this metric enables the student model to achieve state-of-the-art performance in the single-source scenario.

\end{abstract}

\section{Introduction}

In many real-world applications such as sentiment classification \cite{Pang2008}, a model trained on one domain may not work well when directly applied to another domain due to the difference in the data distribution between the domains. At the same time, labeled data in new domains is scarce or non-existent and manual labeling of large amounts of target domain data is expensive. Domain adaptation allows models to reduce the domain discrepancy and adapt to new domains. While fine-tuning is a commonly used method for supervised domain adaptation, there is no cheap equivalent in the unsupervised case as existing Deep Learning-based approaches need to be trained jointly on source and target domain data. This is prohibitive in scenarios with a large number of domains, such as sentiment classification on the plethora of real-world review categories, blog types, or communities \cite{Hamilton2016}. Additionally, re-training a model on source data is unfeasible for evolving domains, such as spam detection where attackers continuously adapt their strategy, scene classification where the scene changes over time \cite{Hoffman2014}, or a conversational agent for a user with a rapidly evolving style, such as a child or second language learner.

Rather than re-training, we would like to be able to leverage our trained model in the source domain to inform the predictions of a new model trained on the target domain. This objective aligns organically with the idea of \emph{Knowledge Distillation} \cite{Bucilua2006, Hinton2015}, which we extend as \emph{Knowledge Adaptation} to the domain adaptation scenario. While Knowledge Distillation concentrates on training a \emph{student} model on the predictions of a (possibly larger) \emph{teacher} model, Knowledge Adaptation focuses on determining \emph{what part} of the teacher's expertise can be trusted and applied to the target domain.

In this context, determining when to trust the teacher is key. This circumstance is paralleled in real-world teacher-student and adviser-advisee relationships: Children learn early on to trust familiar advisers but to moderate that trust depending on the adviser's recent history of accuracy or inaccuracy \cite{Corriveau2009b}, while adults may surround themselves with advisers, e.g. to make a financial investment and gradually learn whose expertise to trust \cite{Johnson2005}.

We demonstrate how domain similarity metrics can be used as a measure of relative trust in a teacher for unsupervised domain adaptation with multiple source domains and show state-of-the-art results for a student model that learns from multiple domain-specific teachers.

When learning from a single teacher in the single-source scenario, using a general measure of domain similarity is inadequate as the student has no other, more relevant teacher to turn to for advice in case its teacher is untrustworthy. To this end, we propose a simple measure, which correlates well with the teacher's accuracy in the target domain and allows the student to gauge the teacher's confidence in its predictions. We demonstrate that by incorporating high-confidence examples selected by this metric in the training process, the student model is able to outperform the state-of-the-art in single-source unsupervised domain adaptation.

Crucially, our models are the first Deep Learning-based models for domain adaptation that perform adaptation without expensive re-training on the source domain data. They are thus able to make use of readily available trained source domain models and are particularly apt for scenarios where domains change or occur in large numbers.

\section{Related work}

\textbf{Distilling knowledge.} \citeauthor{Bucilua2006} \shortcite{Bucilua2006} first proposed a method to compress the knowledge of a source model, which was later improved by \citeauthor{Hinton2015} \shortcite{Hinton2015}. \citeauthor{Romero2015} \shortcite{Romero2015} showed how this method can be adapted to train deep and thin models, while \citeauthor{Kim2016e} \shortcite{Kim2016e} apply the technique to sequence-level models. In addition, \citeauthor{Hu2016} \shortcite{Hu2016} use it to constrain a student model with logic rules. Our goal differs from the previous methods due to the difference in data distributions between source and target data, which necessitates to learn from the teacher's knowledge \emph{only insofar} as it is useful for the target domain. Similar in spirit to Knowledge Distillation is the KL-divergence based objective by \shortcite{Yu2013} \citeauthor{Yu2013} and \cite{Li2014} for adapting an acoustic model and the Adaptive Mixture of Experts model \cite{Nowlan1990}, which also learns which expert to trust for a given example. Both, though, require labeled samples, that are scarce for domain adaptation, while our model is entirely unsupervised.

\textbf{Domain adaptation.} Domain adaptation has a long history of research: \citeauthor{Blitzer2006} \shortcite{Blitzer2006} proposed a structural correspondence learning algorithm. \citeauthor{DaumeIII2007a} \shortcite{DaumeIII2007a} introduced a kernel function that maps source and target domain data to a space that encourages in-domain similarity, while \citeauthor{Pan2010a} \shortcite{Pan2010a} proposed a spectral feature alignment algorithm to align domain-specific words into meaningful clusters, while \citeauthor{Long2015} \shortcite{Long2015} use multi-task learning to avoid negative transfer.

\textbf{Deep learning-based domain adaptation.} Deep learning-based approaches to domain adaptation are more recent and have focused mainly on learning domain-invariant representations: \citeauthor{Glorot2011a} \shortcite{Glorot2011a} first employed stacked Denoising Auto-encoders (SDA) to extract meaningful representations. \citeauthor{Chen2012} \shortcite{Chen2012} in turn extended SDA to marginalized SDA by addressing SDA's high computational cost and lack of scalability to high-dimensional features, while \citeauthor{Zhuang2015} \shortcite{Zhuang2015} proposed to use deep auto-encoders for transfer learning. \cite{Ajakan2015} added a Gradient Reversal Layer that hinders the model's ability to discriminate between domains. Finally, \citeauthor{Zhou2016} \shortcite{Zhou2016} transferred the source examples to the target domain and vice versa using Bi-Transferring Deep Neural Networks, while \citeauthor{Bousmalis2016} \shortcite{Bousmalis2016} propose Domain Separation Networks. All of these approaches, however, require to jointly train the model on source and target data for every new target domain.

\textbf{Domain adaptation from multiple sources.} For domain adaptation from multiple sources, \citeauthor{Mansour2009a} \shortcite{Mansour2009a} proposed a distribution weighted hypothesis with theoretical guarantees. \citeauthor{Duan2009} \shortcite{Duan2009} proposed a method to learn a least-squares SVM classifer by leveraging source classifiers, while \cite{Chattopadhyay2012} assign pseudo-labels to the target data. Finally, \citeauthor{Wu2016a} \shortcite{Wu2016a} exploit general sentiment knowledge and word-level sentiment polarity relations for multi-source domain adaptation.

\section{Knowledge Adaptation}

\subsection{Problem definition}

In the following, we describe domain adaptation within the knowledge adaptation framework: We are provided with one or multiple source domains $\mathcal{D}_{S_i}$ and a target domain $\mathcal{D}_T$. For each of the source domains, we are provided with a teacher model $\mathrm{T_i}$ that was trained on examples $X_{S_i} = \{x_1^{S_i} , \cdots , x_n^{S_i} \}$ and their labels $\{y_1^{S_i}, \cdots, y_1^{S_i} \}$ from $\mathcal{D}_{S_i}$. In the target domain $\mathcal{D}_T$, we only have access to the examples $\{x_1^T , \cdots , x_n^T \}$ without knowledge of their labels. Note that we omit source and target domain indexes in the following for simplicity in cases where examples are unambigous. Our task is now to train a student model $\mathrm{S}$ that performs well on unseen examples from the target domain $\mathcal{D}_T$.

\subsection{Single teacher-student model}

Our teacher and student models are simple multilayer perceptrons (MLP). The basic MLP consists of an input layer, one or multiple intermediate layers, and an output layer. Each intermediate layer $\ell$ learns to embed the output of the previous layer $x$  into a latent representation $h_\ell = f_\ell(W_\ell x + b_\ell)$ where $W_\ell$ and $b_\ell$ are the weights and bias of the $\ell^{th}$ layer, while $f_\ell$ is the activation, typically ReLU $f_l(x) = \max \{0, x \}$ for hidden layers and softmax units $f_l(x) = \mathsf{softmax}(x) = e^x / \sum^{|x|}_{i=1} e^{x_i}$ for the output layer.

In the single source setting, the teacher $\mathrm{T}$ has an output softmax $P_\mathrm{T} = \mathsf{softmax}(z_{\mathrm{T}})$ where $z_\mathrm{T}$ are the logits of the teacher's output layer. $\mathrm{T}$ is trained to minimize the loss $\mathcal{L_\mathrm{T}} = \mathcal{H}(y_i, P_\mathrm{T})$ where $\mathcal{H}$ refers to the cross-entropy and $y_i$ is the label of the $i^{th}$ training example in the source domain $\mathcal{D}_S$.

The student $\mathrm{S}$ similarly models an output probability $P_\mathrm{S} = \mathsf{softmax}(z_{\mathrm{S}})$ where $z_\mathrm{T}$ are the logits of the student's output layer. In the context of knowledge distillation \cite{Hinton2015}, the student $\mathrm{S}$ is trained so that its output $P_\mathrm{S}$ is similar to the teacher's output $P_\mathrm{T}$ and to the true labels. In practice, the output probability of the teacher is smoothed with a temperature $\tau$ to soften the signal and provide more information during training. The same temperature $\tau$ is applied to the output of the student network for the comparison:

\begin{equation}
P_\mathrm{T}^\tau = \mathsf{softmax}(\frac{z_{\mathrm{T}}}{\tau}), \:\:\:\:\: P_\mathrm{S}^\tau = \mathsf{softmax}(\frac{z_{\mathrm{S}}}{\tau}).
\end{equation}

For unsupervised domain adaptation, true labels in the target domain $\mathcal{D_T}$ are not available. Thus the student $\mathrm{S}$ is trained solely to mimic the teacher's softened output with the following loss, which is similar to treating source input modalities as privileged information \cite{Lopez-Paz2016a}:

\begin{equation} \label{eq:unsupervised}
\mathcal{L_\mathrm{S}} = \mathcal{H}(P_\mathrm{T}^\tau, P_\mathrm{S}^\tau).
\end{equation}

\begin{figure*}[!htb]
    \begin{subfigure}{.48\linewidth}
      \centering
         \includegraphics[width=0.9\linewidth]{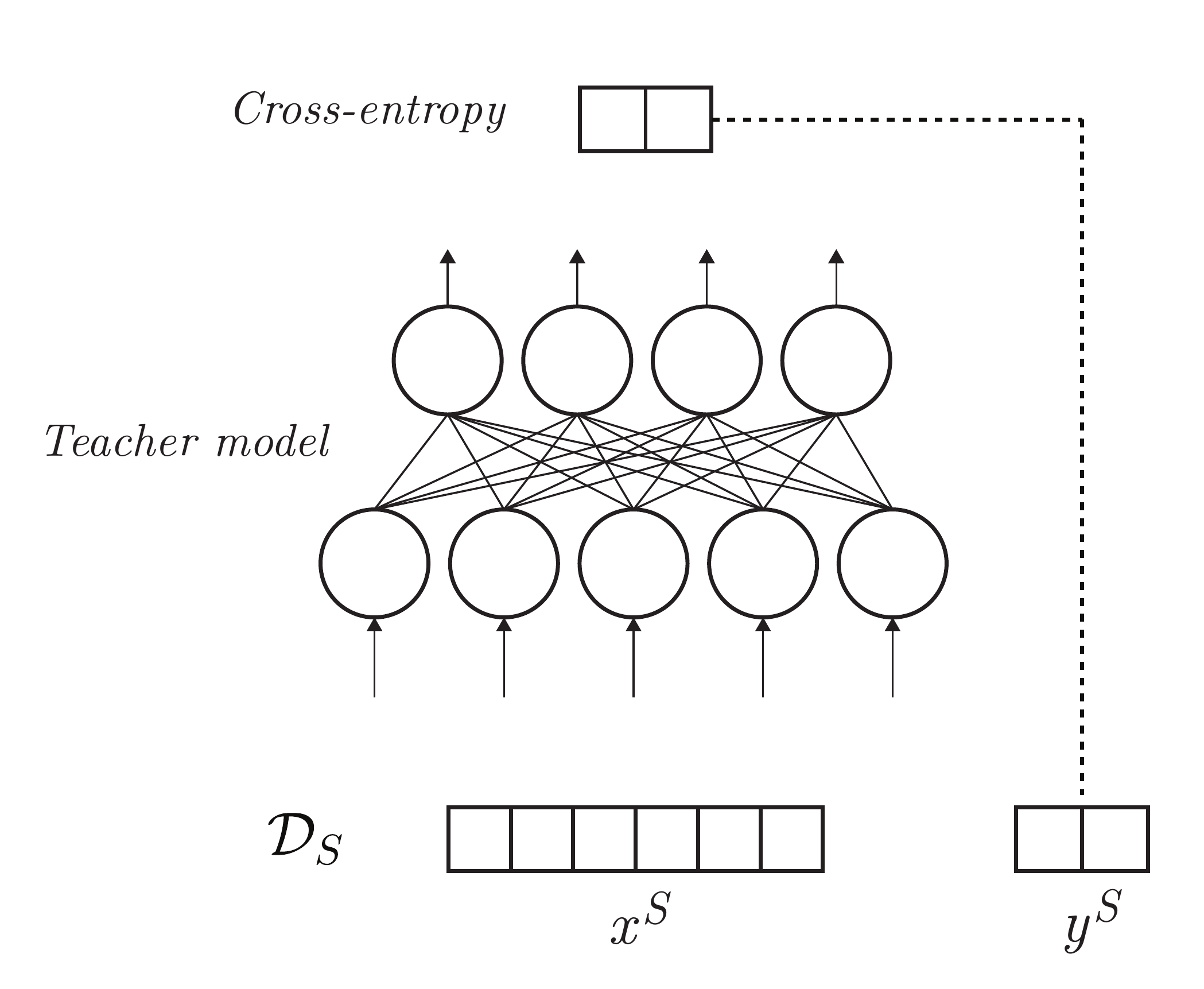}
    \caption{Teacher model} \label{fig:teacher}
    \end{subfigure}%
    \hspace*{0.4cm}
    \begin{subfigure}{.48\linewidth}
      \centering
         \includegraphics[width=0.9\linewidth]{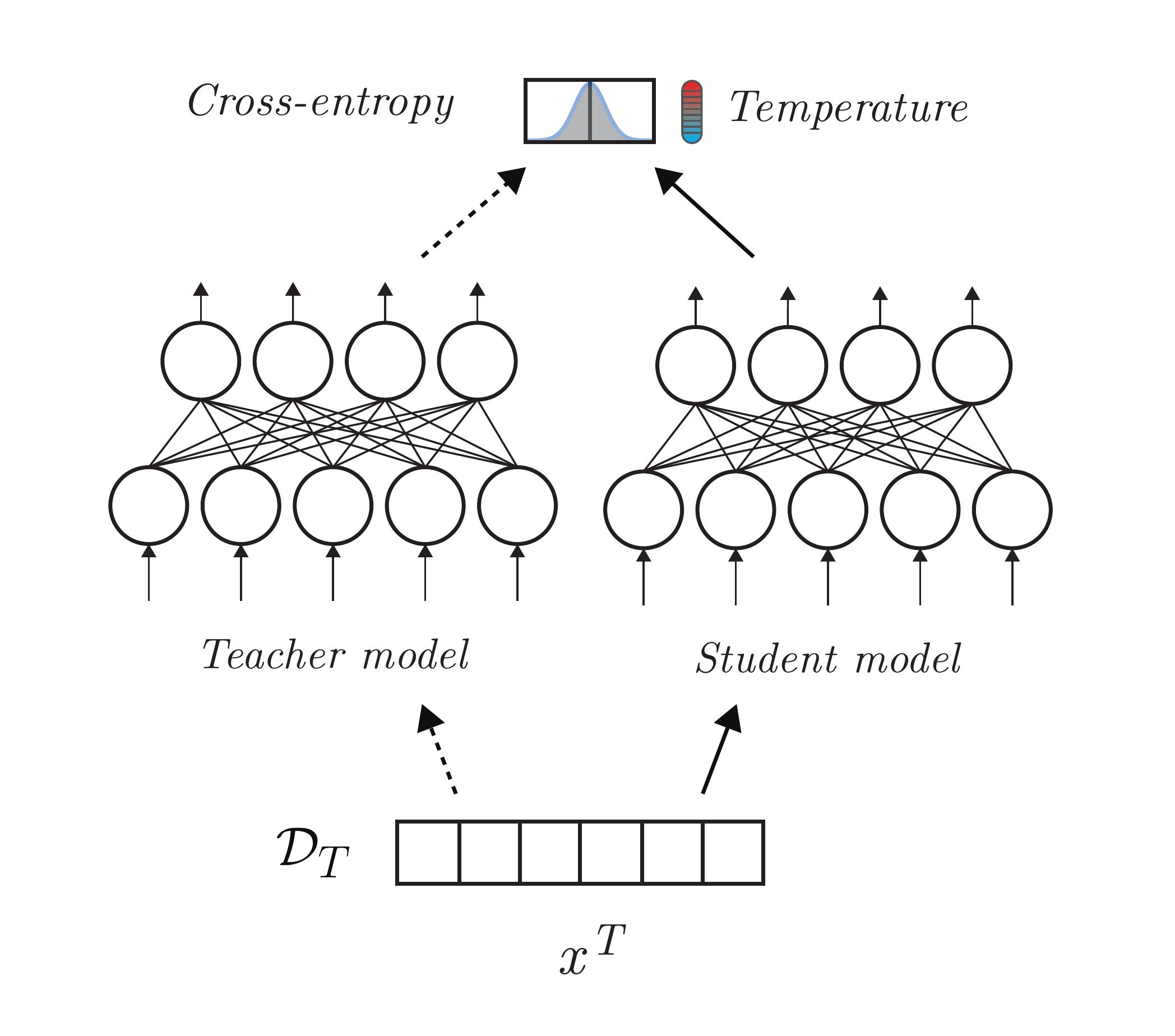}
    \caption{Student model} \label{fig:student}
    \end{subfigure}
    \begin{subfigure}{\linewidth}
      \centering
        \includegraphics[width=0.7\linewidth]{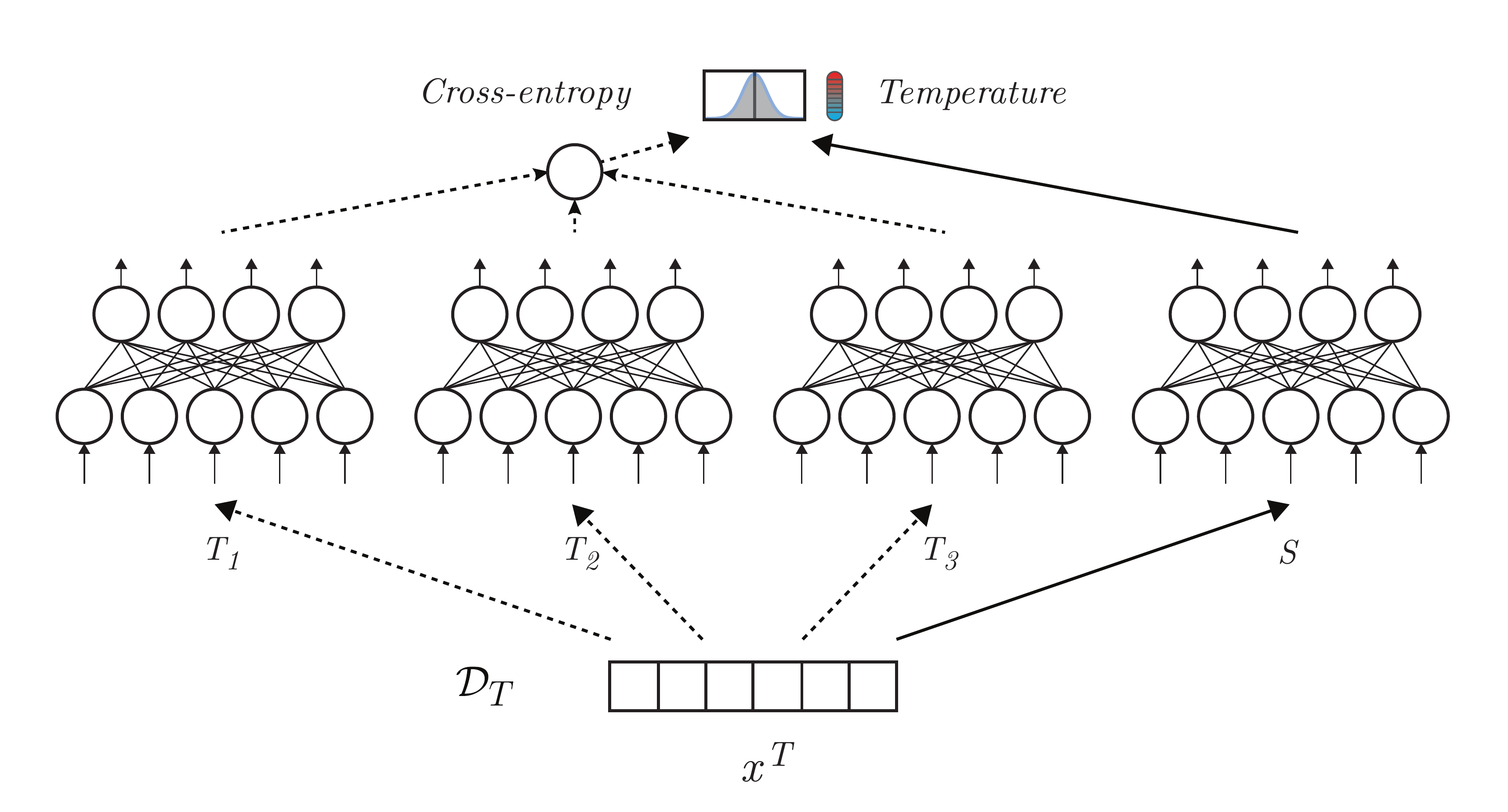}
            \caption{Student model with multiple teachers} \label{fig:multi_teacher_student}
    \end{subfigure}
    \caption{Training procedures for a) the teacher model, b) the student model, and c) the student model with multiple teachers. The teacher is trained on examples $x^S$ and their true labels $y^S$ in the source domain $\mathcal{D}^S$, while the student is trained on the softened predictions of one or multiple teachers of examples $x^T$ in the target domain $\mathcal{D}^T$.}
\label{fig:single-source_results}
\end{figure*}

\subsection{Multiple teacher-student model} \label{sec:multiple_teachers}

The teacher-student paradigm lends itself naturally to the scenario with multiple source domains. Intuitively, the trust that a student should place in a teacher should be proportional to the degree of similarity between the teacher's domain and the student's domain.

To this end, we consider three measures of domain similarity, which have been successfully used in domain adaptation research: Jensen-Shannon divergence \cite{Remus2012} and Renyi divergence \cite{VanAsch2010}, which are both based on Kullback-Leibler divergence and are computed with regard to the domains' term distributions; and Maximum Mean Discrepancy \cite{Tzeng2014}, which we compute with respect to the teacher's latent representation. These measures are computed between the target domain $\mathcal{D}_T$ and every source domain $\mathcal{D}_S$ (additional information with regard to our choice and use of domain similarity measures can be found in the appendix \ref{app:domain_similarity}).

The student model with multiple teachers is then trained to imitate the sum of the teacher's individual predictions weighted with the normalized similarity $sim(\mathcal{D}_S, \mathcal{D}_T)$ of their respective source domain $\mathcal{D}_S$ to the target domain $\mathcal{D}_T$:

\begin{equation}
\mathcal{L}_{MUL} = \mathcal{H}(\sum_{i=1} sim(\mathcal{D}_{S_i}, \mathcal{D}_T) \cdot P_{\mathrm{T}_i}^\tau, P_\mathrm{S}^\tau).
\end{equation}

\subsection{Leveraging a single teacher's knowledge}

General measures of domain similarity are useful in the multi-source setting, where we can rely on multiple teachers and choose to trust one more than the others. In the scenario with a single teacher, it is not helpful to know whether we can trust the teacher \emph{in general}. We rather want a measure that allows us to determine if we can trust the teacher for a \emph{specific example}.

\begin{figure*}[!htb]
    \begin{minipage}{.58\linewidth}
      \centering
         \includegraphics[width=\linewidth]{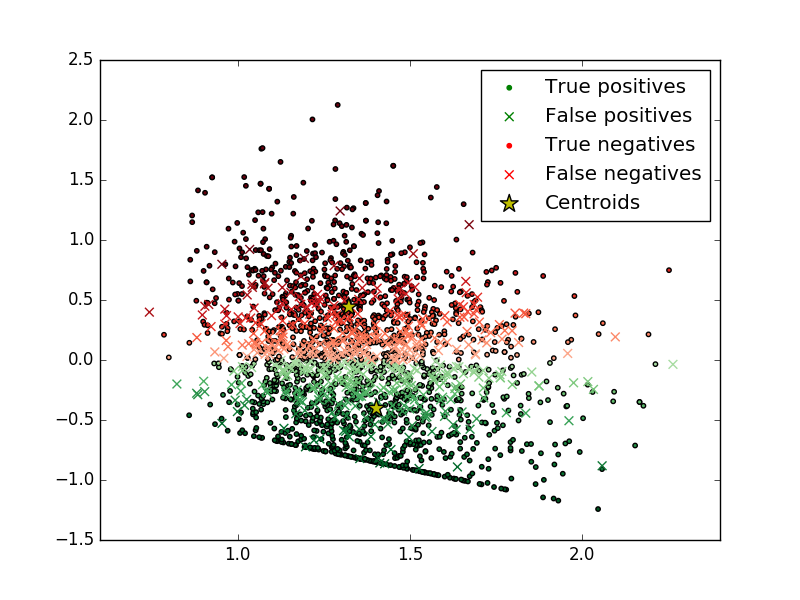}
    \caption{PCA visualization of a teacher's latent representations of target domain examples for the K->D domain pair (see \textsection \ref{sec:data} for details). A darker color reflects a higher MCD value. Best viewed in close-up.}
    \label{fig:pca_mcd}
    \end{minipage}%
    \hspace*{0.4cm}
    \begin{minipage}{.38\linewidth}
      \centering
         \includegraphics[width=\linewidth]{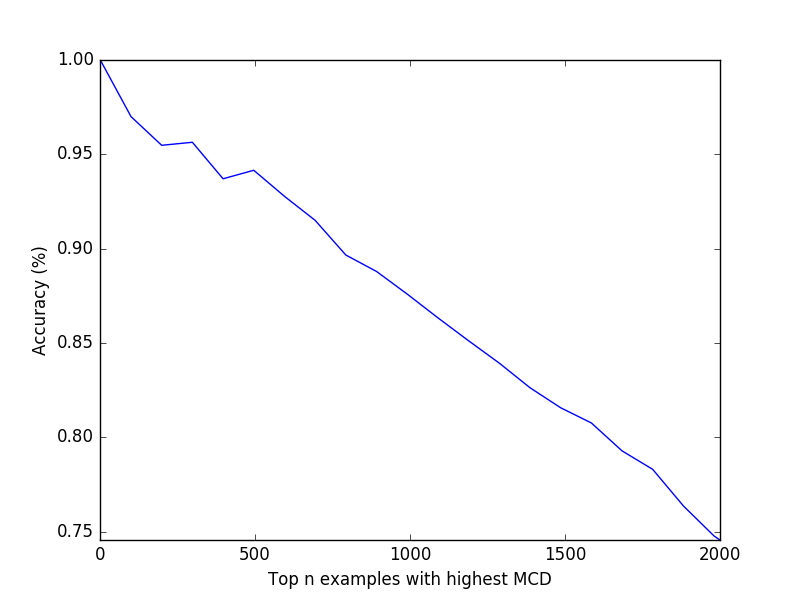}
	\caption{Accuracy of the teacher's predictions on the top $n$ target domain examples with the highest MCD value for the K->D domain pair.}
	\label{fig:acc_mcd}
    \end{minipage}
\end{figure*}

To arrive at such a measure, we revisit the representations the teacher learns from the input data: In order to make accurate predictions, the teacher model learns to separate the representation of different output classes in its hidden representation (we use a one-layer MLP in our experiments as detailed in \textsection \ref{sec:hyperparams}; in deeper networks, this would be an intermediate layer). Even though the teacher model is trained on the source domain, this separation still holds -- albeit with decreased accuracy -- in the target domain. This can be seen in Figure \ref{fig:pca_mcd}, where examples in the target domain that were predicted as positive and negative by the teacher form distinct clusters (refer to \textsection \ref{sec:data} for details with regard to the data and task). Importantly, many of these predictions are incorrect.

As evidenced in Figure \ref{fig:pca_mcd}, incorrect predictions are frequent along the decision boundary and infrequent along the cluster edges, where examples are less ambiguous. More precisely, the accuracy of the teacher's predictions on the target domain is proportional to the absolute difference in similarity of the teacher's representation $h$ with the cluster centroids, which we refer to as Maximum Cluster Difference (MCD) and define as follows:

\begin{equation}
\mathsf{MCD}_h = | \mathsf{cos}(c_p, h) - \mathsf{cos}(c_n, h) |
\end{equation}

where $c_p$ and $c_n$ are the centroids of the positive and negative cluster respectively as predicted by the teacher, i.e. the mean representation of all examples assigned to the cluster by the teacher. Note that while we are focusing on binary classification involving two clusters, the measure is equally applicable to the multi-class setting, as demonstrated in Appendix \ref{app:multiclass_mcd}.

Evidence of the efficacy of this measure for obtaining the trustworthiness of a teacher for an example can be found in the PCA visualization\footnote{A visualization using t-SNE revealed the same cluster. However, PCA showed a clearer decision boundary.} in Figure \ref{fig:pca_mcd}, where incorrect predictions are far less common for (more darkly colored) examples with higher MCD values. Additionally, the MCD score of a target domain example and the accuracy of the teacher's prediction correlate with an average Pearson's $r$ of 0.33 and $p < 0.05$ across all domain pairs of the data described in \textsection \ref{sec:data}. We furthermore plot the teacher's accuracy for the top $n$ target domain examples with the highest MCD values in Figure \ref{fig:acc_mcd}. While the measure becomes less accurate as $n$ increases, it is very accurate for low $n$.

For this reason, rather than weighing all examples with MCD, we propose to add $n$ unlabeled training examples with the highest MCD with their teacher-assigned label as pseudo-supervised examples on which we train the student with the following objective:

\begin{equation} \label{eq:pseudo-supervised}
\mathcal{L_\mathrm{S}} = \mathcal{H}( (1-\lambda) \cdot y_\text{teacher} + \lambda P_\mathrm{T}^\tau, P_\mathrm{S}^\tau)
\end{equation}


\begin{table*}[]
\centering
\begin{tabular}{l c c c c}
 & \textbf{Book} & \textbf{DVD} & \textbf{Electronics} & \textbf{Kitchen} \\
None & 0.7821 & 0.7913 & 0.8181 & \textbf{0.8529} \\
Renyi divergence & 0.7722 & 0.7727 & 0.8133 & 0.8420 \\
Maximum Mean Discrepancy & 0.7811 & 0.7839 & 0.7890 & 0.8273 \\
Jensen-Shannon divergence & \textbf{0.7918} & \textbf{0.7968} & \textbf{0.8203} & 0.8523
\end{tabular}
\caption{Comparison of the impact of different domain similarity measures on the student's performance when used for interpolating the predictions of the source domain teacher models. For the results in each column, the domain in the column header is used as target domain and the remaining three domains are used as source domains.}
\label{tab:domain_similarity}
\end{table*}

where $y_\text{teacher}$ is the indicator array containing $1$ at the index $\mathsf{argmax}(P_\mathrm{T})$ and $0$ at all other indexes, while $\lambda$ determines the contribution of the soft targets. This can be seen as a representation-based variant of instance adaptation \cite{Jiang2007}, which uses MCD as a measure of confidence as it correlates better with teacher accuracy than teacher prediction probability. In practice, we alternate unsupervised training with the objective in equation \ref{eq:unsupervised} and pseudo-supervised training with the objective in equation \ref{eq:pseudo-supervised}, although other curricula are imaginable.

\section{Experiments}

\subsection{Data set} \label{sec:data}

We use the Amazon product reviews sentiment analysis dataset of \citeauthor{Blitzer2006} \shortcite{Blitzer2006}, a common benchmark for domain adaptation. The dataset consists of 4 different domains: Book (B), DVDs (D), Electronics (E) and Kitchen (K). We follow the conventions of past work and evaluate on the binary classification task where reviews with more than 3 stars are considered positive and reviews with 3 stars or fewer are considered negative. Each domains contains 1,000 positive, 1,000 negative, and approximately 4,000 unlabeled reviews. For fairness of comparison, we use the raw bag-of-words unigram/bigram features pre-processed with tf-idf as input \cite{Blitzer2006}.

For single-source adaptation, we replicate the set-up of previous methods and train our teacher models on all 2,000 labeled examples, of which we reserve 200 as dev set. For domain adaptation from multiple sources, we follow the conventions of \citeauthor{Bollegala2011} \shortcite{Bollegala2011} and limit the total number of training examples for all teachers to 1,600, i.e. given three source domains, each teacher is only trained on about 533 labeled samples. We also train a general teacher on the same 1,600 examples of the three domains. In both scenarios, the student is evaluated on all 2,000 labeled samples of the target domain. As we have not found a universally applicable way to optimize hyperparameters or perform early stopping for unsupervised domain adaptation, we choose to use a small number of unlabeled examples as a labeled validation set similar to \cite{Bousmalis2016}.

%
%
%

\subsection{Hyperparameters} \label{sec:hyperparams}

Both student and teacher models are one-layer MLPs with 1,000 hidden dimensions. We use a vocabulary size of 10,000, a temperature of 5, a batch size of 10, and Adam \cite{Kingma2015} as optimizer with a learning rate of 0.001. For every experiment, we report the average of 10 runs.

\subsection{Domain adaptation from multiple sources}

As it is easier for the student to assign trust when learning from multiple teachers, we first conduct experiments on the sentiment analysis benchmark for domain adaptation from multiple sources. For each experiment, one of the four domains is used as the target domain, while the remaining ones are treated as source domains.

\textbf{Domain similarity.} We first evaluate the performance of our student depending on different measures of domain similarity, with which we interpolate the predictions of the teachers. As evidenced in Table \ref{tab:domain_similarity}, Jensen-Shannon divergence generally performs best. We thus use this measure for the remainder of the experiments.

\begin{table*}[]
\centering
\begin{tabular}{l c c c c }
 & \textbf{Book} & \textbf{DVD} & \textbf{Electronics} & \textbf{Kitchen} \\
SCL \cite{Blitzer2006} & 0.7457 & 0.7630 & 0.7893 & 0.8207 \\
SFA \cite{Pan2010a} & 0.7598 & 0.7848 & 0.7808 & 0.8210 \\
SCL-com & 0.7523 & 0.7675 & 0.7918 & 0.8247 \\
SFA-com & 0.7629 & 0.7869 & 0.7864 & 0.8258 \\
SST \cite{Bollegala2011} & 0.7632 & 0.7877 & 0.8363 & 0.8518 \\
IDDIWP \cite{Yoshida2011} & 0.7524 & 0.7732 & 0.8167 & 0.8383 \\
DWHC \cite{Mansour2009a} & 0.7611 & 0.7821 & 0.8312 & 0.8478 \\
DAM \cite{Duan2009} & 0.7563 & 0.7756 & 0.8284 & 0.8419 \\
CP-MDA \cite{Chattopadhyay2012} & 0.7597 & 0.7792 & 0.8331 & 0.8465 \\
SDAMS-SVM \cite{Wu2016a} & 0.7786 & 0.7902 & \textbf{0.8418} & 0.8578 \\
SDAMS-Log \cite{Wu2016a} & 0.7829 & 0.7913 & 0.8406 & 0.8629\\\\
Teacher-only & 0.7565 & 0.7765 & 0.7960 & 0.8210 \\
Student (source teachers) & 0.7918 & 0.7968 & 0.8203 & 0.8523 \\
Student (general teacher) & \textbf{0.8014} & 0.8062 & 0.8365 & \textbf{0.8675}\\
Student (source teachers + general) & 0.8010 & \textbf{0.8088} & 0.8311 & 0.8647
\end{tabular}
\caption{Average results for domain adaptation from multiple sources for the comparison models and ours on the sentiment analysis benchmark. For the results in each column, the domain in the column header is used as target domain and the remaining three domains are used as source domains.}
\label{tab:multi-source_results}
\end{table*}

\textbf{Our models.} For multi-source domain adaptation, we first consider a teacher-only baseline (Teacher-only), where teacher sentiment probabilities are combined, weighted with Jensen-Shannon divergence, and the most likely sentiment is chosen. We further train our student on a) the source domain-specific teachers as detailed in \textsection \ref{sec:multiple_teachers}, b) the general teacher trained on all source domains as described in \textsection \ref{sec:data}, and on c) the combination of source domain and general teachers.

\textbf{Comparison models.} We compare our models against the following methods: domain adaptation with structural correspondence learning (SCL) \cite{Blitzer2006}; domain adaptation based on spectral feature alignment (SFA) \cite{Pan2010a}; adaptations of SCL and SFA via majority voting to the multi-source scenario (SCL-com and SFA-com); cross-domain sentiment classification by constructing a sentiment-sensitive thesaurus (SST) \cite{Bollegala2011}; multiple-domain sentiment analysis by identifying domain dependent/independent word polarity (IDDIWP) \cite{Yoshida2011}; three general-purpose multiple source domain adaptation methods (DWHC, \cite{Mansour2009a}), (DAM, \cite{Duan2009}), (CP-MDA, \cite{Chattopadhyay2012}); cross-domain sentiment classification by transferring sentiment along a sentiment graph with hinge loss and logistic loss respectively (SDAMS-SVM and SDAMS-Log) \cite{Wu2016a}. Numbers are taken from \citeauthor{Wu2016a} \shortcite{Wu2016a}.

\textbf{Results.} All results are depicted in Table \ref{tab:multi-source_results}. Evaluating the combination of the source teacher models directly on the target domain (Teacher-only) produces the worst results, which underscores the need for methods that allow adaptation to the target domain. Training the student model on the soft targets of the teachers allows us to improve upon the teacher-only baseline significantly, which demonstrates the appropriateness of the teacher-student paradigm to the domain adaptation scenario. The student model outperforms comparison methods that rely on source model predictions by combining \cite{Mansour2009a} or predicting \cite{Duan2009} them. This showcases the usefulness of learning from soft targets in the domain adaptation scenario. Training on a general teacher model as well as on a combination of the general teacher and the source domain teachers allows us to improve results even further.
 Both models improve over existing approaches to domain adaptation from multiple sources and outperform approaches that rely on sentiment analysis-specific information \cite{Wu2016a} in all but the electronics domain.

\subsection{Single-source domain adaptation}

We additionally evaluate the ability of the student to only learn from a single teacher. This scenario is more challenging as the student cannot consider other teachers that might provide more relevant predictions. For each target domain, each of the three other domains is used as source domain, yielding 12 domain pairs.

\textbf{Our models.}  On these domain pairs, we firstly evaluate our student-teacher (TS) model. For training a model that incorporates high-confidence predictions of the teacher (TS-MCD), we cross-validate the interpolation parameter $\lambda$ in equation \ref{eq:pseudo-supervised} and the number of examples with the highest MCD scores $n$. We find that a low $\lambda$ (around 0.2) generally yields the best results in the domain adaptation setting, as the high-confidence predictions are helpful to guide the student's learning during training. Additionally, using the top 500 unlabeled target domain examples with the highest MCD scores for pseudo-supervised training of the student produces the best results.

\begin{figure*}[!htb]
    \begin{minipage}{.48\linewidth}
      \centering
         \includegraphics[width=\linewidth]{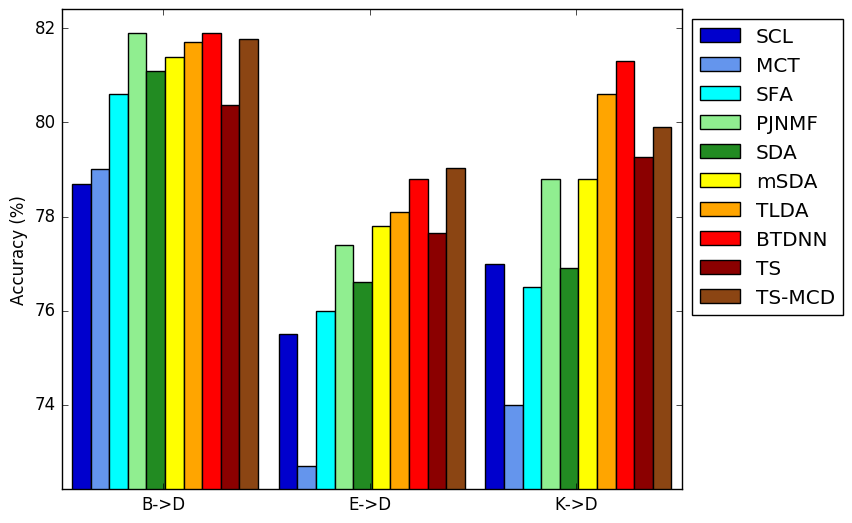}
    \end{minipage}%
    \hspace*{0.4cm}
    \begin{minipage}{.48\linewidth}
      \centering
         \includegraphics[width=\linewidth]{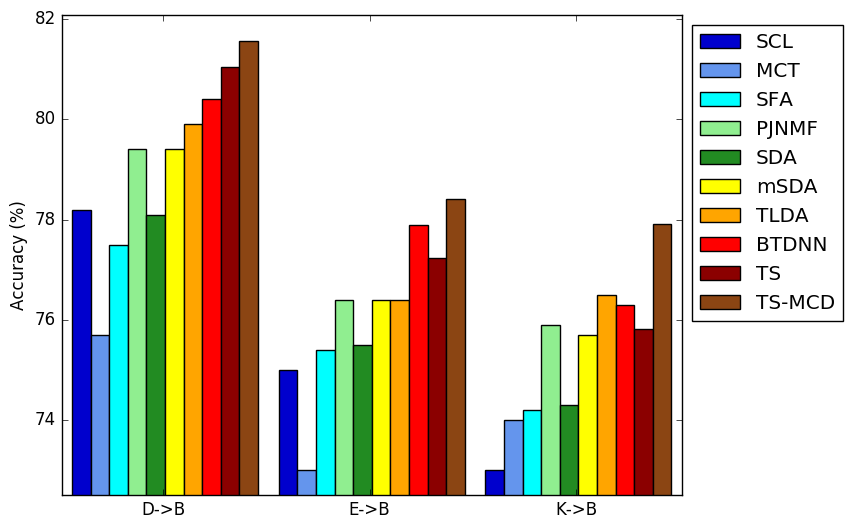}
    \end{minipage}
    \begin{minipage}{.48\linewidth}
      \centering
        \includegraphics[width=\linewidth]{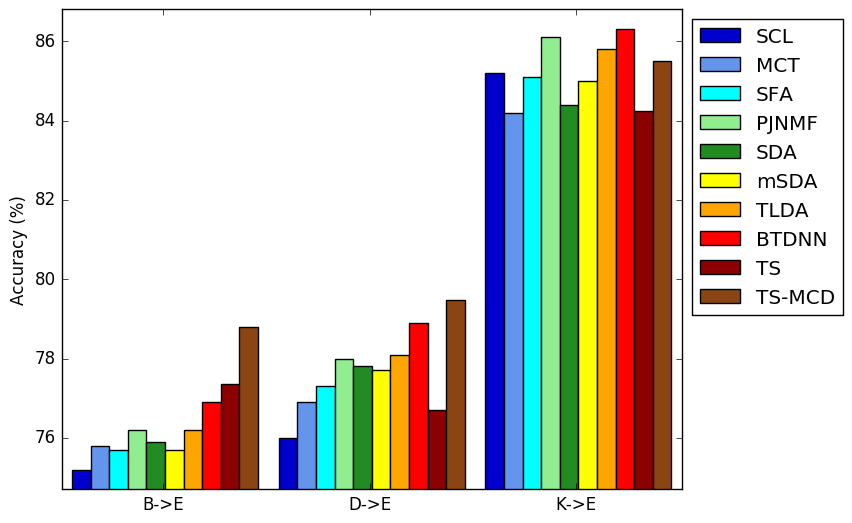}
    \end{minipage}%
    \hspace*{0.4cm}
    \begin{minipage}{.48\linewidth}
      \centering
        \includegraphics[width=\linewidth]{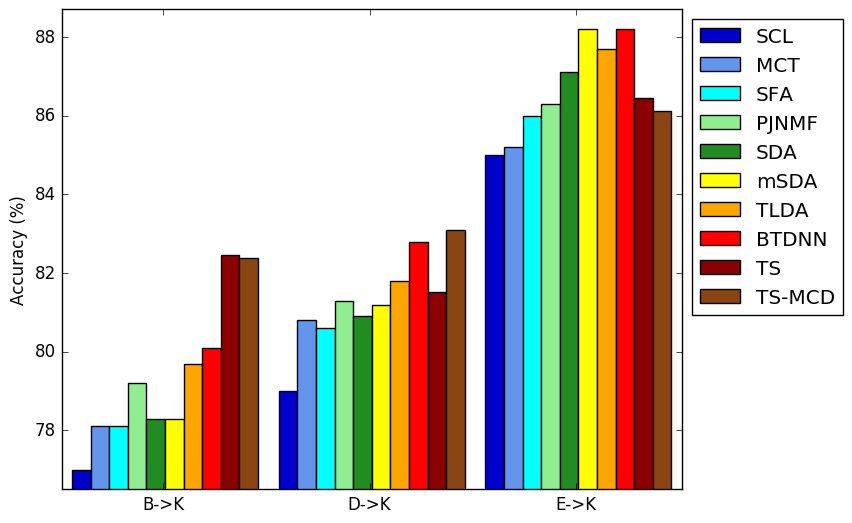}
    \end{minipage}
    \caption{Average results for single-source domain adaptation for the comparison models and our models on the sentiment analysis benchmark. B: Book. D: DVD. E: Electronics. K: Kitchen.}
\label{fig:single-source_results}
\end{figure*}

\textbf{Comparison models.} $\:$ For the single-source case, we similarly compare against SCL \cite{Blitzer2006} and SFA \cite{Pan2010a}, as well as against multi-label consensus training (MCT), which combines base classifiers trained with SCL \cite{Li2008} and against an approach that links heterogeneous input features with points via non-negative matrix factorization (PJNMF) \cite{Zhou2015c}. We additionally compare against the following deep learning-based approaches: stacked denoising auto-encoders (SDA) \cite{Glorot2011a}; marginalized SDA (mSDA) \cite{Chen2012}; transfer learning with deep auto-encoders (TLDA) \cite{Zhuang2015}; and bi-transferring deep neural networks (BTDNN) \cite{Zhou2016}.

\textbf{Results.} The results can be seen in Figure \ref{fig:single-source_results}. The student trained on the source domain teacher (TS) achieves convincing results and outperforms the state-of-the-art on three domain pairs -- twice with the Book domain as source domain, showing that knowledge acquired from the Book domain might perhaps be more easily transferable to a student model. For many domain pairs, the student still falls significantly short compared to the performance of the state-of-the-art, which highlights that solely relying on a single teacher's predictions is insufficient to bridge the discrepancy between the domains. Instead, additional methods are necessary to provide evidence for the student when to trust the teacher's predictions. Leveraging the teacher's knowledge by incorporating high-confidence examples selected by MCD into the training (TS-MCD) improves the performance of the student in almost all cases significantly. This allows the student to outperform the state-of-the-art on 8 out of 12 domain pairs without expensive joint training on source and target data and with the sole dependence of a single model trained on the source domain, which is typically readily available.

\section{Conclusion}

In this work, we have proposed Knowledge Adaptation, an extension of the Knowledge Distillation idea to the domain adaptation scenario. This method -- in contrast to prevalent domain adaptation methods -- is able to perform adaptation without re-training. We firstly demonstrated the benefit of this paradigm by showing that a student model that takes into account the predictions of multiple teachers and their domain similarities is able to outperform the state-of-the-art for multi-source unsupervised domain adaptation on a standard sentiment analysis benchmark. We additionally introduced a simple measure to gauge the trustworthiness of a single teacher and showed how this measure can be used to achieve state-of-the-art results on 8 out of 12 domain pairs for single-source unsupervised domain adaptation.

\bibliography{knowledge_adaptation_acl}
\bibliographystyle{acl_natbib}

\appendix

\section{Appendix}

\subsection{Domain similarity measures} \label{app:domain_similarity}

We use three measures of domain similarity in our experiments: Jensen-Shannon divergence, Renyi divergence, and Maximum Mean Discrepancy (MMD).

Jensen-Shannon divergence is a smoothed, symmetric variant of KL divergence. The Jensen-Shannon divergence between two different probability distributions $P$ and $Q$ can be written as:

\begin{equation}
D_{JS}(P||Q) = \frac{1}{2} [D_{KL}(P||M) + D_{KL}(Q||M)]
\end{equation}

where $M = \frac{1}{2} (P + Q)$, i.e. the average distribution of $P$ and $Q$, and $D_{KL}$ is the KL divergence:

\begin{equation}
D_{KL}(P||Q) = \sum_{i=1}^n p_i \frac{p_i}{q_i}.
\end{equation}

Renyi divergence similarly generalizes KL divergence by assigning different weights to the probability distributions of the source and target domain and is defined as follows: 

\begin{equation}
D_R(P||Q) = \frac{1}{\alpha-1} \: \text{log}(\sum_{i=1}^n \frac{p_i^\alpha}{q_i^{\alpha-1}}).
\end{equation}

If $\alpha=1$, Renyi divergence reduces to KL divergence. In our experiments, we set $\alpha=0.99$ following \cite{VanAsch2010}.

These domain similarity measures are typically based on the term distributions of the source and target domains, i.e. the probability distribution $P$ of a domain is the term distribution $t \in \mathcal{R}^{|V| \times 1}$ where $t_i$ is the relative probability of word $w_i$ appearing in the domain and $|V|$ is the size of the vocabulary of the domain. The intuition behind using term distributions is that similar domains usually have more terms in common than dissimilar domains. While term distributions are efficient to compute and have proven effective in previous work \cite{VanAsch2010,Wu2016a}, they only capture shallow occurrence statistics.

Another form of similarity metrics such as MMD are based on representations. MMD measures the distance between a source and target distribution with respect to a particular representation $\phi$. The MMD between the source data $X_S$ and the target data $X_T$ is defined as follows:

\begin{equation}
\begin{split}
MMD(X_S, X_T) = \| & \frac{1}{ |X_S| } \sum_{x_S \in X_S} \phi(x_S) - \\ 
& \frac{1}{ |X_T| } \sum_{x_T \in X_T} \phi(x_T) \|.
\end{split}
\end{equation}

The representation $\phi$ is usually obtained by embedding the source data and target data in a Reproducing Kernel Hilbert Space via a specifically chosen kernel, e.g. \cite{Bousmalis2016} use a linear combination of RBF kernels. Similarly to \cite{Tzeng2014}, we use the hidden representation of a neural network as basis for $\phi$, as we are interested in how well the teacher's representation captures difference in domain.

In our experiments, MMD does not outperform the more traditional term distribution-based similarity measure, which we attribute to two reasons: 1) Due to the limited amount of data, our teacher model is not deep enough to capture the difference in domain in its single hidden layer; \cite{Tzeng2014} in contrast identify the fully-connected layer $fc7$ in the AlexNet architecture as the layer minimizing MMD. 2) The teacher is only trained on the source domain data. Its representation is thus not sensitive to detect the domain shift to the target domain. Training a separate model to minimize MMD alleviates this, but incurs additional computational costs and requires retraining on the source data during adaptation, which we set out to avoid to enable efficient adaptation.

Another commonly used measure of domain similarity is $\mathcal{A}$-distance. \cite{Ben-David2007} show that computing the $\mathcal{A}$-distance between two domains reduces to minimizing the empirical risk of a classifier that tries to discriminate between the examples in those domains. Previous work \cite{Blitzer2007} uses the Huber loss and a linear classifier for computing the $\mathcal{A}$-distance. In our experiments, $\mathcal{A}$-distance did not outperform Jensen-Shannon divergence, while its reliance on training a classifier is a downside in our scenario with multiple or changing target domains, where we would prefer more efficient measures of domain similarity.

\subsection{Multi-class MCD} \label{app:multiclass_mcd}

Maximum cluster difference can be easily extended to the multi-class setting. For $n$ classes, we compute $n$ cluster centroids for the clusters whose members have been assigned the same class by the model. We then create a set $C$ containing all $n(n-1)/2$ unique pairs of cluster centroids. Finally, we compute the sum of pair-wise differences of the model's representation $h$ with regard to the cluster centroid pairs:

\begin{equation}
\mathsf{MCD}_\text{multi} = \sum_{c_1, c_2 \in C} | \mathsf{cos}(c_1, h) - \mathsf{cos}(c_2, h) |.
\end{equation}

\end{document}